\documentclass{article}
\usepackage{ijcai25}

\usepackage{times}
\usepackage{soul}
\usepackage{url}
\usepackage[hidelinks]{hyperref}
\usepackage[utf8]{inputenc}
\usepackage[small]{caption}
\usepackage{graphicx}
\usepackage{amsmath}
\usepackage{amsthm}
\usepackage{booktabs}
\usepackage{algorithm}
\usepackage{algorithmic}
\usepackage[switch]{lineno}

\usepackage{subcaption}
\usepackage{bbold}
\usepackage{amssymb}
\usepackage{array}
\usepackage{bbding}
\usepackage{multirow}

\title{Prompt-Free Conditional Diffusion for Multi-object Image Augmentation}

\author{
Haoyu Wang$^1$
\and
Lei Zhang$^{1}$\footnotemark[2]\and
Wei Wei$^1$\and
Chen Ding$^2$\And
Yanning Zhang$^1$\\
\affiliations
$^1$Northwestern Polytechnical University\\
$^2$Xi'an University of Posts \& Telecommunications\\
\emails
wanghaoyunwpu@mail.nwpu.edu.cn,\\
\{nwpuzhanglei, weiweinwpu, ynzhang\}@nwpu.edu.cn,
dingchen@xupt.edu.cn
}

\begin{document}

\maketitle

\renewcommand{\thefootnote}{\fnsymbol{footnote}}
\footnotetext[2]{Corresponding author.}

\begin{abstract}
Diffusion models has underpinned much recent advances of dataset augmentation in various computer vision tasks. However, when involving generating multi-object images as real scenarios, most existing methods either rely entirely on text condition, resulting in a deviation between the generated objects and the original data, or rely too much on the original images, resulting in a lack of diversity in the generated images, which is of limited help to downstream tasks.
To mitigate both problems with one stone, we propose a prompt-free conditional diffusion framework for multi-object image augmentation. Specifically, we introduce a local-global semantic fusion strategy to extract semantics from images to replace text, and inject knowledge into the diffusion model through LoRA to alleviate the category deviation between the original model and the target dataset. In addition, we design a reward model based counting loss to assist the traditional reconstruction loss for model training. By constraining the object counts of each category instead of pixel-by-pixel constraints, bridging the quantity deviation between the generated data and the original data while improving the diversity of the generated data.
Experimental results demonstrate the superiority of the proposed method over several representative state-of-the-art baselines and showcase strong downstream task gain and out-of-domain generalization capabilities. Code is available at \href{https://github.com/00why00/PFCD}{here}.
\end{abstract}

\section{Introduction}
In the past decade, deep neural networks have achieved a surge of success in a wide range of computer vision tasks ~\cite{resnet,vit,CLIP}. One key premise for such success lies on the collection of large-scale training images. However, in real scenarios even for a specific single task, amassing sufficient images to establish a dataset is often prohibitively costly and laboriously time-intensive, e.g., imageNet~\cite{ImageNet} for image classification. For this problem, a promising solution proves to be image augmentation with generative models~\cite{GANaug},  which aims at randomly generating extensive synthetic images based on a few manually collected images to rapidly establish a dataset. Following this idea, various effective image generative models~\cite{generate4classification,generate4UDA} have been proposed successively. Among them, profiting from the powerful generative capacities, diffusion models have been paid increasing attention to image generation and augmentation. Usually, given some prompts related to the scene content, the diffusion model can directly generate a high-quality image with such a content.

\begin{figure*}[!ht]
\centering
\includegraphics[width=1.0\linewidth]{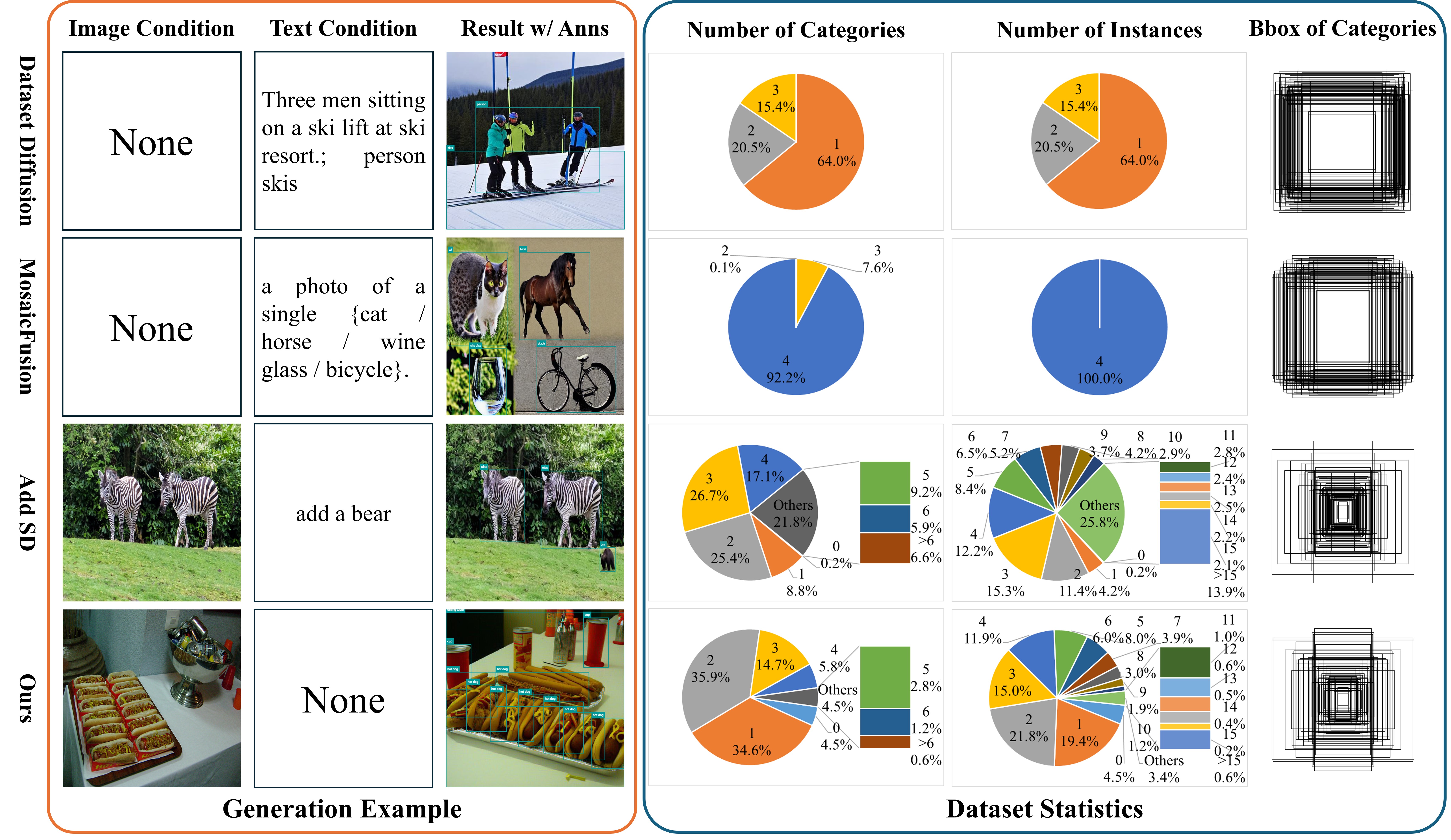}
\caption{Comparison with state-of-the-art image augmentation methods. Dataset Diffusion decrease in object amount with low annotation quality. MosaicFusion generate counterfactual images and objects are similar in size. Add SD cannot change the background of the image and have low variation in layout. Our method generates a large number of objects while ensuring image diversity.}
\label{fig: dataset stat}
\end{figure*}

In real-world applications, generating multi-object images with complex spatial relationships is crucial. Although some recent progress have been made for multi-object image generation, due to much increased generation difficulty, most of these methods suffer from obvious limitations. As shown in Fig. \ref{fig: dataset stat}, several existing methods \cite{DatasetDM,DatasetDiffusion} use category names or image captions as conditions to inputs into the pre-trained diffusion model to generate images, and use attention maps to extract image labels. However, it is difficult to generate a large number of objects using only text prompts, and the quality of labels generated using attention maps is poor when objects overlap. Although some methods \cite{InstanceDiffusion,ParaDiffusion} use stronger guidance, such as layout or paragraph, to improve the quality of generated images, these methods are difficult to scale. Some methods \cite{X-Paste,MosaicFusion} solve this problem by decomposing the multi-object image generation task. They first generate single-object images and their corresponding labels through a pre-trained diffusion model, and then use data augmentation to synthesize multi-object images. Although the number of objects and label quality are increased, artificial facts are often generated, which reduces the reality of the image.

In addition, the above methods tend to pursue training-free and directly use simple text prompts containing category names to generate data, which leads to deviations in style, size, etc. between the generated images and the original data. Furthermore, some methods \cite{Gen2Det,AddSD} try to replace or add objects to the original images through image editing. Although this makes the augementated image as realistic as possible, the amount of information added is limited because the layout, background and most of the objects in the image have not changed.

To fill this gap, we propose a prompt-free conditional diffusion framework, aims to reduce the category and quantity deviations from the original data while improving the diversity of generated images. Inspired by image variation task \cite{UnCLIP,VersatileDiffusion,PromptFreeDiffusion}, our framework utilizes a single multi-object image instead of text prompts as the condition of the diffusion model to reduce the category bias brought by text descriptions. More importantly, to better extract and inject the multi-object information into the diffusion procedure, we propose a local-global semantic fusion strategy that utilizes the pre-trained CLIP \cite{CLIP} model to separate extract the semantic knowledge within the whole condition image as well as its local crop. On the other hand, to further control the object amount as well as the layout diversity in the generated image, we further propose a reward model based counting loss to explicitly restrict the amount of objects in each category in the generated image, while imposing no any constraint on their spatial layout. By doing this, the proposed model is able to randomly generate high-quality images with the same number of objects in each category as the condition image or even more but showing different layouts, thus guaranteeing the variety of image augmentation. Experimental results demonstrate the superiority of the proposed method over several representative state-of-the-art baselines and showcase good downstream task gains and out-of-domain generalization capabilities.

In summary, this study mainly contributes in four aspects:

\begin{enumerate}
\item We propose a prompt-free conditional diffusion framework for multi-object image augmentation. By changing the text condition to a novel local-global semantic fusion strategy, which enables appropriate extracting the multi-object information from the condition image and injecting it into the diffusion model for image generation.
\item We design a reward model based counting loss to constrain the number of objects in each category of generated images, which improves the diversity of images.
\item We contribute new state-of-the-art performance of both downstream tasks and generated quality on MS-COCO dataset in terms of multi-object image augmentation.
\end{enumerate}

\section{Related Work}

\begin{figure*}[htbp]
\centering
\includegraphics[width=0.8\textwidth]{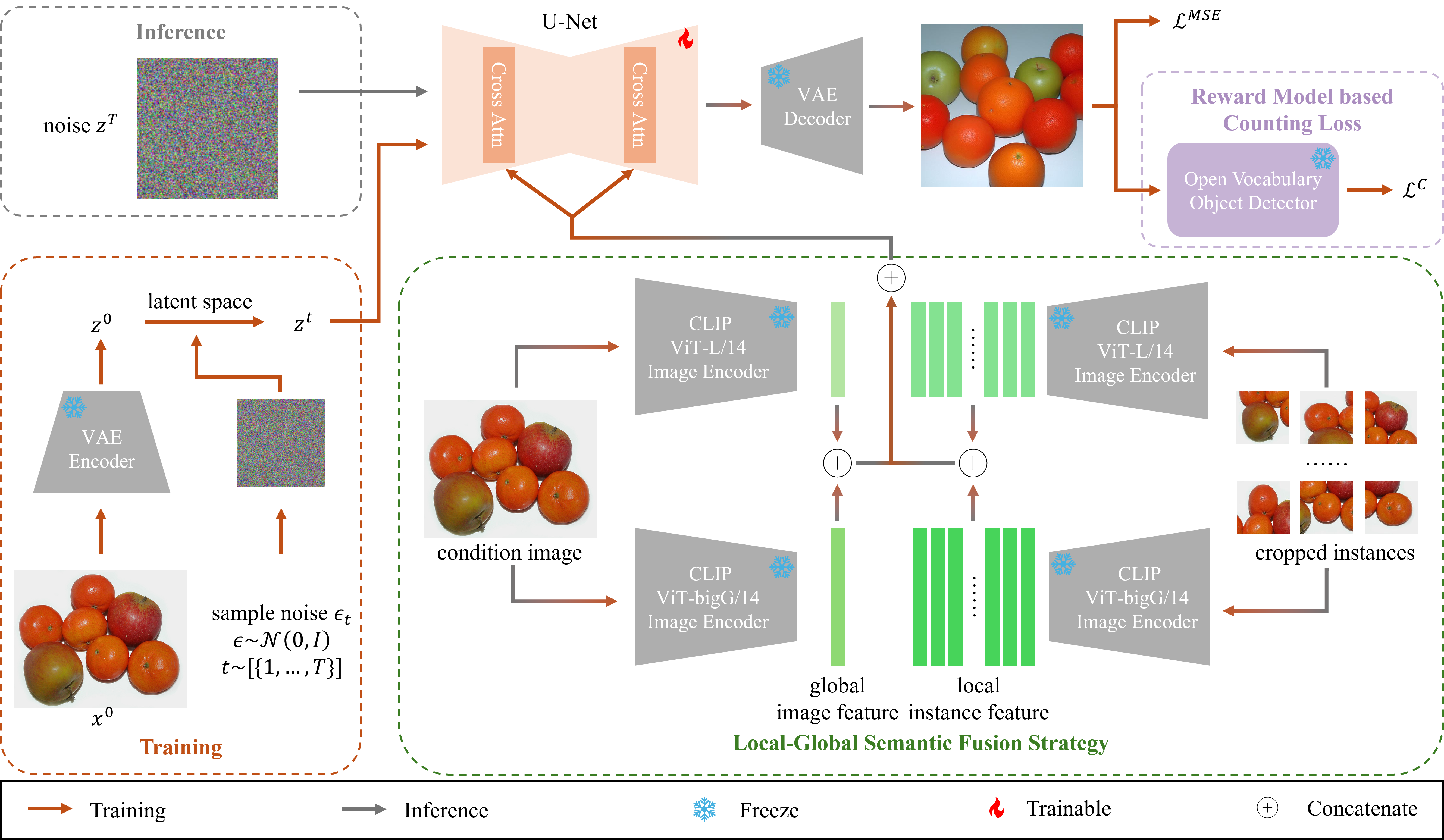}
\caption{Overview of the proposed prompt-free conditional diffusion framework. We introduce a local-global semantic fusion strategy to generate images with local instance categories and global semantics comparable to the condition image. We also introduce a reward model based counting loss to ensure that the number of objects in each category of the image do not decrease.}
\label{fig: overview}
\end{figure*}

\subsection{Text-to-Image Diffusion Models}

Driven by multi-modal technology, text-to-image diffusion models exhibit formidable capabilities in image generation. GLIDE \cite{GLIDE} uses a cascade architecture and classifier-free guidance \cite{CFG} for image generation based on pre-trained language models. DALL-E2 \cite{UnCLIP} adopts a multi-stage model, using CLIP \cite{CLIP} text encoder to encode text and images. Imagen \cite{Imagen} uses multiple text encoders to improve sample fidelity and text-image alignment. The latent diffusion model \cite{StableDiffusion} significantly reduces computational overhead by transferring the diffusion process from the image to a low-dimensional feature space. On its basis, exemplar-based methods \cite{ControlNet++} achieve refined control of generated images under the guidance of text by introducing structural information as input, such as mask, edge, pose, etc. Subject-driven image generation methods \cite{DreamBooth,TextualInversion} realize the customized generation of specific objects under the guidance of several target images and relevant text prompts. In contrast to the above techniques, the goal of our framework does not require specifying locations or customization of individuals for each instance, but to generate factual images with comparable object amounts and diverse layouts.

\subsection{Image Variation}

Given an image, image variation aims to generate an image with similar styles or semantics. Currently, there is no unified paradigm for image variation tasks. DALL-E2 \cite{UnCLIP} uses the alignment characteristics of the CLIP image and text encoder to encode input images to achieve image variation. ControlNet \cite{ControlNet} controls the generation of the diffusion model by adding an additional network structure based on the latent diffusion model. 
Its reference-only version achieves variation images by splicing the original attention layer of the diffusion model with the attention layer of the control network. Versatile Diffusion \cite{VersatileDiffusion} designs a multi-stream multimodal latent diffusion model framework and supports the diversified generation of a single image stream. Prompt-Free Diffusion \cite{PromptFreeDiffusion} replaces the text encoder with a semantic context encoder to learn the features of the input image and diversify it. Compared with the method proposed in this paper, the above method performs diversification on the entire image, and its diversified connotation often includes multiple information such as content, style, and color, which cannot guarantee that the instance of the generated image is consistent with the original image.

\section{Methodology}


\subsection{Problem Formulation}

Despite the availability of excellent annotation tools such as SAM 2 \cite{SAM2} and Grounding DINO \cite{GroundingDINO}, the diversified generation of large-scale multi-object images remains a problem that needs to be solved. In multi-object dataset augmentation, consider a collection of $ N $ samples, denoted as $ \mathcal{D}=\{(x_i, y_i), i=1,..., N\} $, where $ x_i=\{(o_j,c_j), j=1,..., N_i^c\} $, represents an input image containing $ N_i^c $ categories, and for each category $ c_j $, it contains $ o_j $ objects. $ y_i=\{(b_k,c_k),k=1,..., N_i^o\} $ denotes the category $ c_k $ and the structured box annotations $ b_k $ of $ N_i^o $ objects, and $ \sum_{j=1}^{N_i^c}o_j=N_i^o $. The goal of the task is to generate a set of enhanced images $ \mathcal{D}^*=\{x_i^*,i=1,..., N\} $ with the same number of input samples, where $ x_i^*=\{(o_l,c_l),l=1,..., N_i^{c*}\} $, requiring that for each category $ c_j $ in the input image $ x_i $, a $ c_l $ can be found in the augmented image $ x_i^* $ corresponding to it, and the count of it $ o_l \geq o_j $.

\subsection{Overall Architecture}

As shown in Fig. \ref{fig: overview}, the proposed framework consists of two parts: a local-global semantic fusion strategy and a reward model based counting loss.

During the forward diffusion process, the pre-trained latent diffusion model first uses the encoder $ E $ to compress the input image $ x_i^0 = x_i \in \mathbb{R}^{H \times W \times 3} $ into a latent representation $ z_i^0 \in \mathbb{R}^{h \times w \times d} $, while the decoder $ D $ can transform the latent representation into pixel space, i.e., $ D(z_i^0) \approx x_i^0 $, where $ \frac{H}{h} = \frac{W}{w} = 8 $ and $ d = 4 $. Then the noise $ \epsilon_t $ is sampled from the Gaussian distribution and added to it, where $ t $ is a time step sampled from the uniform distribution. Finally, a DDPM is trained in the latent space based on the image condition $ p_i^{img} $ using the MSE loss and our proposed counting loss to recover $ z_i^0 $ from the Gaussian distribution, where the MSE loss $ \mathcal{L}_i^{MSE} $ is:
\begin{equation}
    \mathcal{L}_i^{MSE} = \mathbb{E}_{z \sim E(x), y, \epsilon \sim \mathcal{N}(0,1), t}[||\epsilon - \epsilon_\theta(z_i^t,t,\mathcal{C}(p_i^{img}))||_2^2],
\end{equation}
where $ z_i^t $ is the noise feature of time step $ t $, $ \epsilon_\theta $ is the noise prediction network, which takes $ z_i^t $ as input and predicts the sampled Gaussian noise guided by the time step $ t $ and the conditional feature $ \mathcal{C}(p_i^{img}) $, where $ \mathcal{C} $ is our proposed local-global semantic fusion module.

For the reverse diffusion process, the model directly samples noise in the latent space and uses the trained noise prediction network to gradually denoise it according to the conditional features to obtain the final image.

\subsection{Local-Global Semantic Fusion}

Numerous papers \cite{T2I1,T2I2,T2I3} point out that the text-to-image diffusion model often fails to generate images that accurately match the text prompt, especially when the prompt contains information such as multiple categories or counts. In addition, since most current multi-object generation methods pursue training-free and directly use text prompts to generate images, the generated category distribution is offset from the target dataset distribution due to the inherent bias of the generation model. To address the challenges of category bias introduced by text-based prompts, we replace textual prompts with image-based conditions for diffusion models. Using images as input conditions better captures the category distribution of the target dataset, reducing deviations and improving the fidelity of the generated data.

In order to adapt the latent diffusion model from text-guided image generation to image-guided image generation, we use the image encoder $ E_{img} $ pre-trained together with the original text encoder $ E_{text} $ using paired text-image data to encode the image condition, so that the obtained conditional features remain in the same feature space without fine-tuning all the parameters of the diffusion model.

\begin{algorithm}[t]
\caption{Counting Loss}
\label{alg: counting loss}
\textbf{Input}: denoised image $ x_i^* $, open vocabulary object detector $ \mathcal{D}_{OV} $, number of categories $ N_i^c $, text prompt $ S_{i} $, class count list $ L_i^{count} $, class index list $ L_i^{index} $, counting loss step $ \gamma $, counting loss threshold $ \tau $
\begin{algorithmic}[1]
\IF{training steps larger than $ \gamma $}
\STATE Let $ logits_{i} \gets \mathcal{D}_{OV}(x_i^*, S_{i}) $.
\FOR{$ j \gets 1,...,N_i^c $}
\IF{$ L_i^{index}[j] $ is an integer}
\STATE Let $ s_i^j \gets logits_{i}[L_i^{index}[j]] $
\ELSE
\FOR{idx in $ L_i^{index}[j] $}
\STATE Let $ s_i^j \gets $ concatenate all $ logits_{i}[idx] $
\ENDFOR
\ENDIF
\STATE Calculate $ \mathcal{L}_{i}^{j} $ use Eq. \ref{eq: class loss}
\ENDFOR
\STATE Calculate $ \mathcal{L}_i^C $ use Eq. \ref{eq: norm}
\STATE \textbf{return} $ \mathcal{L}_i^C $
\ENDIF
\end{algorithmic}
\end{algorithm}

The original text encoder uses hidden states of text conditions to capture the semantic relationship between the text context:
\begin{equation}
    \mathcal{C}(p_{text}) = E_{text}(T(p_{text})),
\end{equation}
where $ \mathcal{C}(p_{text})\in \mathbb{R}^{bs \times seq \times emb} $ is the output conditional feature, $ bs $ is the batch size of text condition $ p_{text} $, $ seq $ is the sequence length, $ emb $ is the feature dimension, and $ T $ is the tokenizer. In order to further clarify the instance that needs to be enhanced, we crop it from the image, merge it with the original image and input it into the image encoder to extract features:
\begin{equation}
    p_i^{img} = \{x_i, Crop(x_i, b_i, pad)\},
\end{equation}
where $ p_i^{img} $ is the local-global semantic fusion condition, and $ Crop(\cdot) $ uses the bounding box information $ b_i $ of the global image $ x_i $ to crop the local instance to be augmented. In order to better understand the image context, we use hyperparameter $ pad $ to control the pixel of outward cropping.

Through the above operations, we express the information that is difficult to control with text, such as count and category, through batched local-global image information, highlighting its importance in the condition. To reduce computational complexity, we only need all the features of the original image, and for each cropped image, we only need its $ [CLS] $ feature:
\begin{equation}
\mathcal{C}(p_i^{img})=E_{img}(P(p_i^{img},M))
\end{equation}
where $ \mathcal{C}(p_i^{img})\in \mathbb{R}^{bs \times (1 + M) \times emb} $, $ P $ is the image processor that processes the image condition for batch training. Specifically, for each image condition $ p_i^{img} $, the image processor fixes its cropped instances to $ M $. When the number of instances is less than $ M $, it is expanded with zero tensors, otherwise $ M $ instances are randomly selected for training. We set $ M $ to 9, which significantly reduces the computational complexity compared to the text condition while ensuring the semantics of most objects.

\begin{figure*}[htb]
\centering
\includegraphics[width=0.87\textwidth]{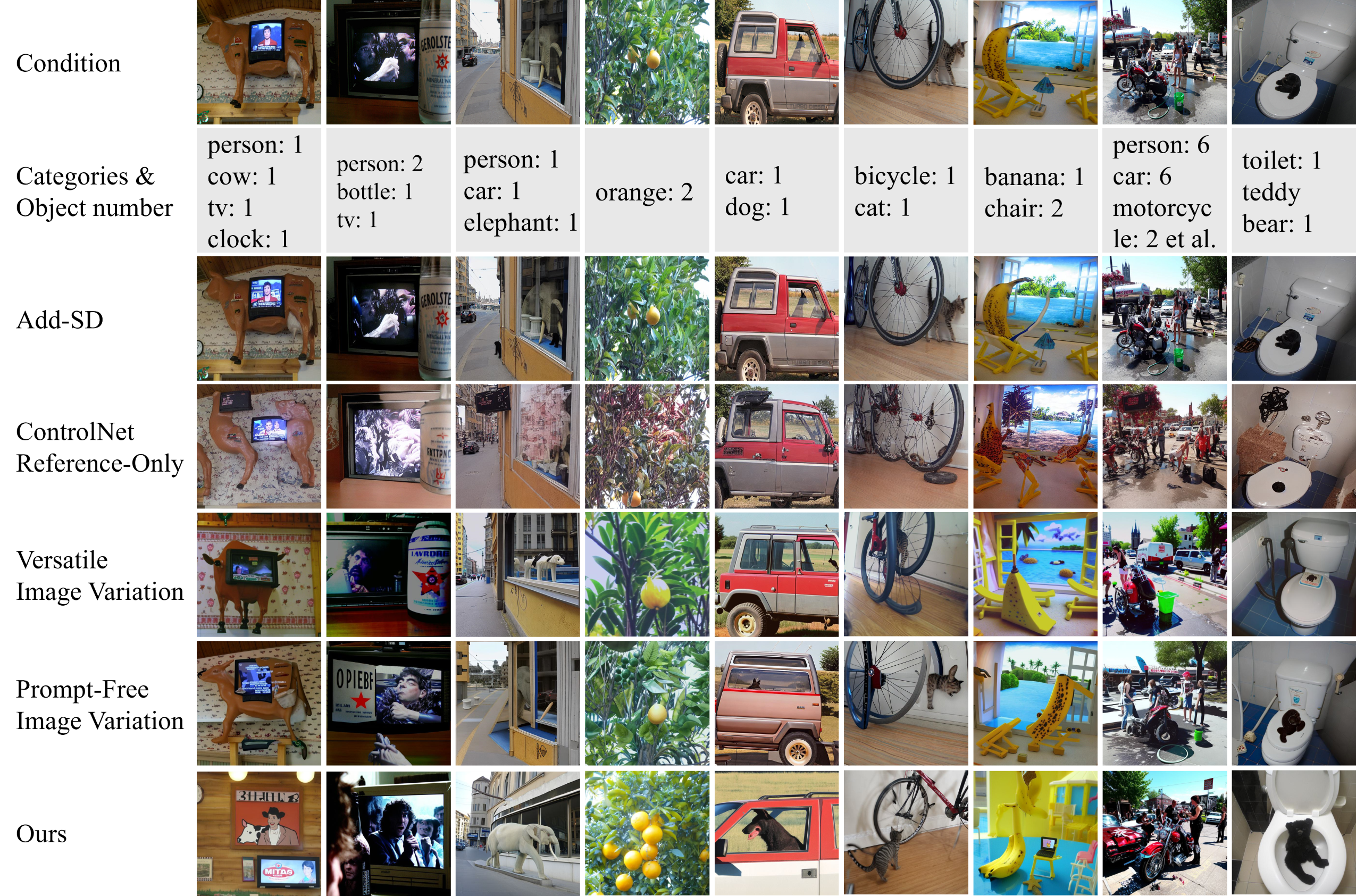}
\caption{Qualitative comparison. We compare with Dataset Diffusion w/SDXL and SDXL img2img, ControlNet Reference-Only, Versatile Image Variation and Prompt-Free Image Variation on COCO 2017 validation set. Our method is superior to other methods in terms of sufficient objects, realism, and layout diversity. Better viewed with zoom-in.}
\label{fig: main-result}
\end{figure*}

\subsection{Reward Model Based Counting Loss}

With the proposed local-global semantic fusion strategy, we can improve the fidelity of the generated image. To further ensure that the object amounts do not degrade, we propose a reward model based counting loss. Specifically, for the input image $ x_i $, we obtain the image $ x_i^* $ by one-step denoising during training:
\begin{equation}
    x_i^* = \frac{1}{\sqrt{\alpha_t}}(x_i^t - \frac{1-\alpha_t}{\sqrt{1-\overline\alpha_t}} \epsilon_\theta(x_i^t, t)) + \sigma_t \mathbf{z},
\end{equation}
where $ \epsilon_t $ is the noise prediction network, $ \alpha_t $, $ \overline\alpha_t $, $ \sigma_t $ are the hyperparameters defined by DDPM, and $ \mathbf{z} \sim \mathcal{N}(0,1) $ is used to adjust the signal-to-noise ratio.

Then we use the image annotations to construct supervision information. The proposed counting loss focuses solely on category counts, rather than bounding box positions, to promote diverse layout generation. This design ensures that object counts match the desired distribution without imposing rigid spatial constraints, enhancing both flexibility and diversity in the generated images. For the $ N_i^c $ categories contained in the image, we first count the number of objects in each category and obtain a one-to-one corresponding category name list $ L_i^{class} = \{name(c_j), j=1,...,N_i^c\} $ and count list $ L_i^{count} = \{len(o_j), j=1,...,N_i^c\} $, where $ name(\cdot) $ is used to get the category name, and $ len(\cdot) $ is a function of counting numbers. Then we connect each name in $ L_i^{class} $ with a period to construct the text prompt $ S_i $ of the reward model. Since some category names have more than one word, we also record the index list $ L_i^{index} $ of each category in $ S_i $ to obtain the result of the reward model.

Finally, we use the pre-trained open vocabulary object detector as the reward model to detect the categories in the image according to $ S_{i} $. For detection result of $ c_j $, we take the highest confidence sample based on input image and calculate the loss according to the threshold hyperparameter $ \tau $:
\begin{equation}
\label{eq: class loss}
\mathcal{L}_{i}^{j} = \sum_{L_i^{count}[j]} ReLU(\tau - topk(s_{i}^{j}, k=L_i^{count}[j])),
\end{equation}
where $ s_i^j $ is the confidence result of in category $ c_j $  of image $ x_i $ detected by the reward model. And the final counting loss $ \mathcal{L}_i^C $ of image $ x_i $ is:
\begin{equation}
\label{eq: norm}
\mathcal{L}_i^C = \frac{\sum_{j=1}^{N_i^c}(\mathcal{L}_{i}^{j})}{\sum_{j=1}^{N_i^c}(L_{i}^{count}[j])}.
\end{equation}

The hyperparameter $\gamma$ determines the training step at which counting loss begins to take effect. This avoids noisy gradients during early training stages when the denoised images may still contain significant noise. The calculation method of counting loss is outlined in Algorithm \ref{alg: counting loss}. The overall training loss of the proposed framework can be formulated as:
\begin{equation}
\label{eq: total loss}
\mathcal{L} = \sum_i (\mathcal{L}_i^{MSE} + \lambda \mathcal{L}_i^C),
\end{equation}
where $ \lambda $ is a hyperparameter for adjusting the loss weight.

\section{Experiments}

\subsection{Experimental Setups}

\subsubsection{Datasets}

We validate the proposed framework and comparison methods on the MS-COCO \cite{COCO} dataset, a relatively complex object detection dataset containing 80 categories, with an average of 7.7 objects per image. We use $ train2017 $ containing 118K images to train the proposed method and generate images for downstream task evaluation, and use the COCO validation set $ val2017 $ consisting of 5K images for generation quality evaluation.

\subsubsection{Implementation Details}

We use Stable Diffusion XL \cite{SDXL} and Grounding DINO \cite{GroundingDINO} as LDM and reward model respectively. We fine-tune the model using LoRA \cite{LoRA} at 512 $ \times $ 512 resolution, we set the learning rate to 1e-4, total batch size to 32, and train on two RTX 3090 GPUs using the AdamW \cite{AdamW} optimizer with constant scheduler. For training images, we use center crop and random flip as data augmentation. In the inference stage, we use the Euler scheduler with 50 steps for generation.

\subsubsection{Metrics}

In addition to the qualitative results, we use multiple quantitative indicators to evaluate our proposed method from various dimensions. For downstream task evaluation, we use mAP (mean Average Precision) and AP50 to evaluate the generated data, and for generation quality evaluation, we use the widely used Frechet Inception Distance (FID) \cite{FrechetInceptionDistance} to evaluate the fidelity of the generated images. In addition, to evaluate the diversity of the generated images, we calculate the diversity score (DS) by comparing the LPIPS \cite{LPIPS} metric of paired images. Finally, to evaluate the object amounts of the generated images, we designed an instance quantity score (IQS) that detects the instance quantity of each category under multiple confidence settings using the pre-trained YOLOv8m \cite{YOLOv8} and compares it with the original images. Algorithm is shown in Appendix. 

\subsection{Comparison Methods}

We compare the proposed method with the state-of-the-art multi-object image augmentation methods Dataset Diffusion \cite{DatasetDiffusion}, Mosaic Fusion \cite{MosaicFusion}, Add SD \cite{AddSD}, and image variation methods ControlNet Reference Only \cite{ControlNet}, Versatile Diffusion \cite{VersatileDiffusion}, and Prompt-Free Diffusion \cite{PromptFreeDiffusion}. We use the pre-trained model of the above methods to generate images under its default parameters.


\subsection{Downstream Task Evaluation}

To verify the effectiveness of the proposed method, we use the generated data to train downstream detection and segmentation models and compare the metrics of the validation set to test the ability of image augmentation. Specifically, we use the training set of the COCO dataset to generate 10k data for each method and mix it with the original training set to train Mask RCNN\cite{maskrcnn}. For Dataset Diffusion, we use the pre-trained model to perform instance segmentation on its semantic labels. For Add SD, image variation methods, and our method, we use Grounding DINO\cite{GroundingDINO} and SAM\cite{SAM} to generate annotations.

Tab. \ref{tab: downstream tasks} shows the performance indicators of all methods on the validation set. Our method achieves the state-of-the-art performance on both detection models. These results demonstrate the effectiveness of the proposed method and provide promising results for the further application of generative models in detection tasks.

\begin{table}[tb]
\centering
\small
\begin{tabular}{lcccc}
\toprule
\multirow{2}{*}{\textbf{Task}} & \multicolumn{2}{c}{\textbf{bbox}} & \multicolumn{2}{c}{\textbf{mask}} \\
 & mAP & AP50 & mAP & AP50 \\
\toprule
train2017 & 38.65 & 59.48 & 35.24 & 56.32 \\
\midrule
Dataset Diffusion & 38.34 & 59.09 & 35.12 & 56.12 \\ 
MosaicFusion & 38.74 & 59.60 & 35.15 & 56.45 \\
Add SD & 37.88 & 58.67 & - & - \\
\midrule
ControlNet & 38.75 & 59.62 & 35.33 & 56.39 \\
Versatile & 38.80 & 59.50 & 35.36 & 56.50 \\
Prompt-Free & 38.68 & 59.23 & 35.15 & 56.21 \\
\midrule
Ours & \textbf{39.04} & \textbf{59.86} & \textbf{35.43} & \textbf{56.73} \\
\bottomrule
\end{tabular}
\caption{Performance comparison of downstream task evaluations across state-of-the-art methods.}
\label{tab: downstream tasks}
\end{table}

\subsection{Generation Quality Evaluation}

To further verify the generation quality of the model, we use the validation set of the COCO dataset to evaluate the generated images. Specifically, we use each image in the validation set as a condition for image augmentation and calculate the fidelity, diversity score, and instance quantity score of all images. Since Dataset Diffusion \cite{DatasetDiffusion} and MosaciFusion \cite{MosaicFusion} do not support image-based augmentation, we do not compare with them here.

\subsubsection{Qualitative Evaluation}

Fig. \ref{fig: main-result} shows the visualization results on some challenging samples. 
Add SD\cite{AddSD} does not always successfully add targets. ControlNet\cite{ControlNet} cannot understand the semantics of the input image and loses the object of interest after image augmentation. The diversity of the images generated by Versatile Diffusion\cite{VersatileDiffusion} and Prompt Free Diffusion\cite{PromptFreeDiffusion} is not good. The layout of the image is almost the same as the original image, and there will be problems with missing targets and even counterfactual images. Compared with the above methods, our method achieves the best results in the balance of layout diversity, number of generated objects, and consistency with facts.

\begin{table}[b]
\centering
\small
\begin{tabular}{lccc}
\toprule
\textbf{Methods} & \textbf{FID} $ \downarrow $ & \textbf{DS} $ \uparrow $ & \textbf{IQS} $ \uparrow $ \\
\toprule
val2017 & - & - & 45.02 \\
\midrule
Add SD\cite{AddSD} & \textbf{6.90} & 0.19 & \textbf{32.55} \\
\midrule
ControlNet\cite{ControlNet} & 25.50 & 0.64 & 15.91 \\
Versatile\cite{VersatileDiffusion} & 19.01 & \underline{0.65} & 24.64 \\
PFD\cite{PromptFreeDiffusion} & 22.39 & 0.62 & 20.23 \\
\midrule
Ours & \underline{18.59} & \textbf{0.71} & \underline{29.17} \\
\bottomrule
\end{tabular}
\caption{Quantitative comparison with state-of-the-art methods. $ \uparrow $ means higher is better, $ \downarrow $ means lower is better. All generated images are evaluated at 512 $ \times $ 512 resolution.}
\label{tab: main-result}
\end{table}

\subsubsection{Quantitative Evaluation}
As demonstrated in Tab. \ref{tab: main-result}, val2017 represents the results of the original val dataset. The proposed method achieves the best or suboptimal results across FID, DS and IQS metrics, which proves the effectiveness of our method. It is worth noting that although Add SD \cite{AddSD} is superior to the proposed method in terms of fidelity through image editing, its diversity score is greatly reduced, and the instance quantity score is even lower than the original dataset. The proposed method has optimal performance in terms of the balance of fidelity, diversity and instance quantity.

\begin{figure}[t]
\centering
\includegraphics[width=1.0\columnwidth]{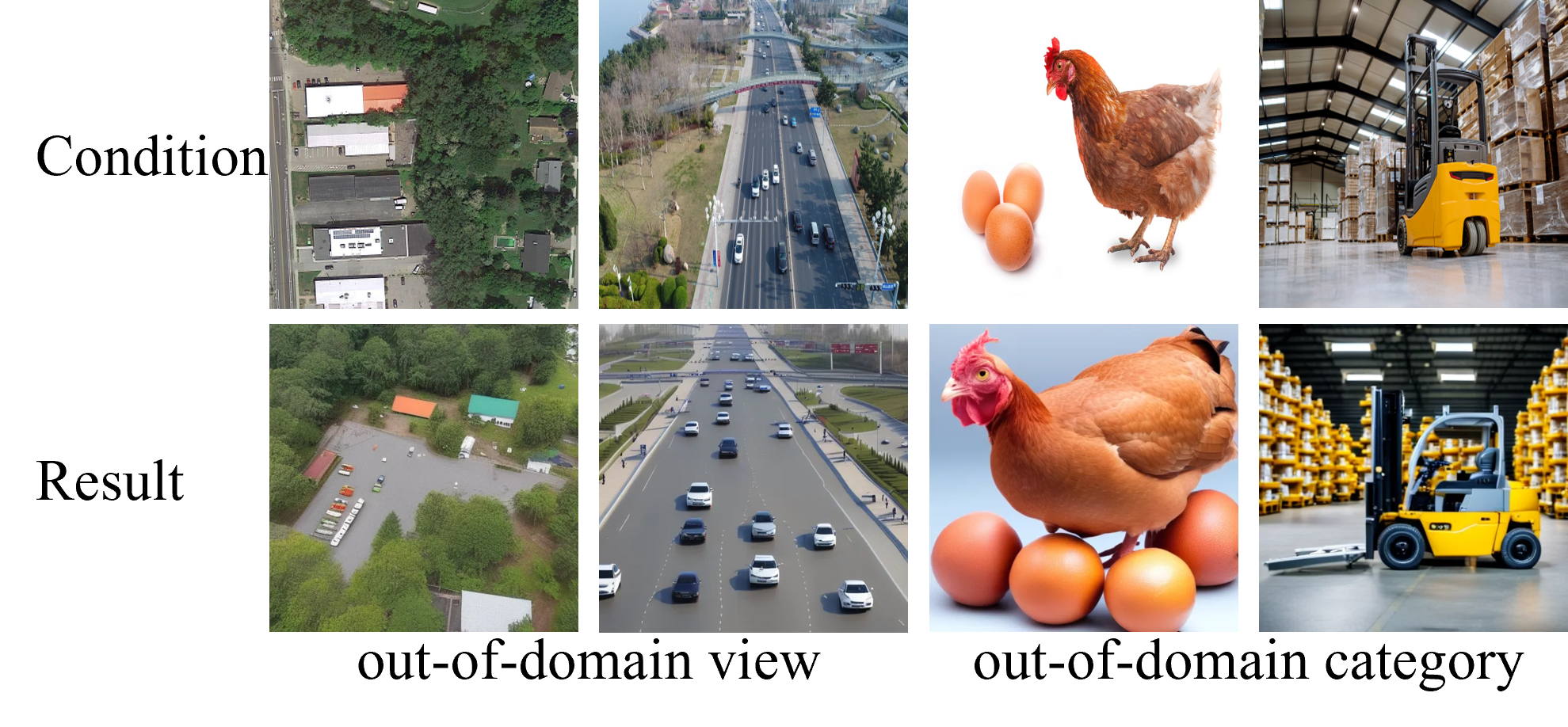}
\caption{Out-of-domain experimental results under two settings.}
\label{fig: ood result}
\end{figure}

\begin{figure}[t]
\centering
\includegraphics[width=1.0\linewidth]{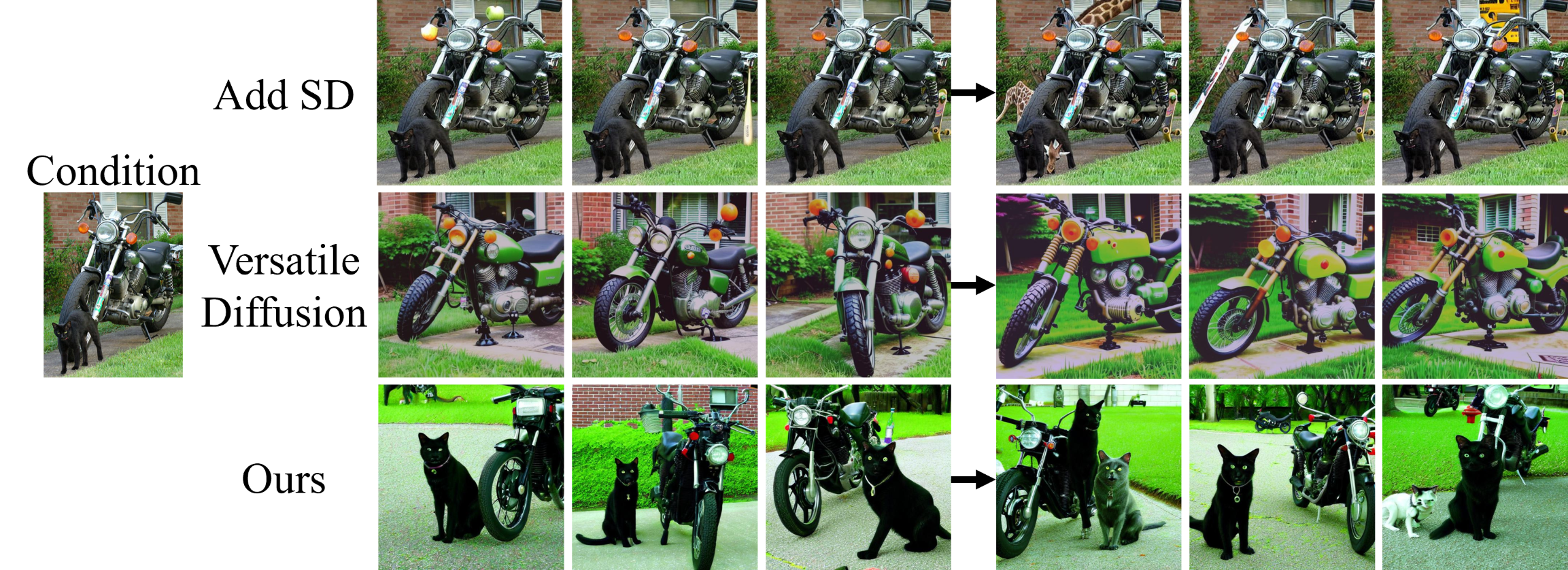}
\caption{Recurrent generation for a given condition}
\label{fig: variation main}
\end{figure}

\subsubsection{Out-of-domain Evaluation}

We also conducted experiments on the out-of-domain generalization ability of the model, including two settings. The cross-view setting is shown in the first two columns of Fig. \ref{fig: ood result}. We experiment with satellite and drone remote sensing images. Our method can correctly understand the semantics in the input image and perform cross-view image augmentation on it. The cross-category setting is shown in the last two columns of Fig. \ref{fig: ood result}. We test categories such as chickens, eggs, forklift, et al. that are not in the COCO dataset. Our method can understand the semantics of unseen categories through only images and generate diverse images.

\subsubsection{Multiple Random Generation \& Recurrent Generation}

As shown in Fig. \ref{fig: variation main}, we verified the effects of different methods on multiple augmentations of a single image and further augmentations of the augmented image. Our method achieved the best layout diversity. More results can be found in the supplementary materials.

\subsection{Ablation Study}

\subsubsection{Effectiveness of Each Component}

To further verify the effectiveness of our proposed components, we conducted a series of ablation studies. These studies mainly focus on two key components of our model: the local-global semantic fusion strategy and the reward model based counting loss. We use SDXL img2img as our baseline. As shown in Tab. \ref{tab: method-ablation}, after using semantic fusion module to replace the original text condition module, the fidelity and diversity of the model have been significantly improved. Similarly, after adding counting loss, the instance quantity score of the model was further improved.

\begin{table}[!t]
\centering
\small
\begin{tabular}{lccc}
\toprule
\textbf{Method} &  \textbf{FID} $ \downarrow $ & \textbf{DS} $ \uparrow $ & \textbf{IQS} $ \uparrow $ \\
\toprule
Baseline & 29.49 & 0.36 & 25.67 \\
+ Semantic Fusion & 20.43 & 0.68 & 27.98 \\
+ Counting Loss & \textbf{18.59} & \textbf{0.71} & \textbf{29.17} \\
\bottomrule
\end{tabular}
\caption{Ablations of local-global semantic fusion strategy (SF) and reward model based counting loss (CL).}
\label{tab: method-ablation}
\end{table}

\begin{table}[!t]
\centering
\small
\begin{tabular}{lccc}
\toprule
\textbf{Method} &  \textbf{FID} $ \downarrow $ & \textbf{DS} $ \uparrow $ & \textbf{IQS} $ \uparrow $ \\
\toprule
Image Only & 19.27 & 0.70 & 27.34 \\
w/ Category Name & 19.02 & 0.70 & 28.22 \\
w/ Content Image & \textbf{18.59} & \textbf{0.71} & \textbf{29.17} \\
w/ Both  & 18.85 & \textbf{0.71} & 28.90 \\
\bottomrule
\end{tabular}
\caption{Ablations of different conditions.}
\label{tab: condition-ablation}
\end{table}

\begin{table}[!t]
\centering
\small
\begin{tabular}{lccc}
\toprule
\textbf{Method} &  \textbf{FID} $ \downarrow $ & \textbf{DS} $ \uparrow $ & \textbf{IQS} $ \uparrow $ \\
\toprule
Random Crop & \textbf{18.42} & 0.71 & 25.99 \\
Grounding DINO & 19.24 & \textbf{0.72} & 27.22 \\
Ground Truth & 18.59 & 0.71 & \textbf{29.17} \\
\bottomrule
\end{tabular}
\caption{Ablations of different inference methods.}
\label{tab: inference-ablation}
\end{table}

\subsubsection{Analysis of Different Conditions}

We also conducted ablation experiments under different conditions. As shown in Tab. \ref{tab: condition-ablation}, although both category name and content conditions can improve the performance of the model, the category name is a subset of the content, and using only the content can enable the model to achieve higher performance.

\subsubsection{Analysis of Different Inference Methods}

In practical applications, we cannot always obtain the ground truth of image annotations. So we use random crop and Grounding DINO detection results as contents for inference. As shown in Tab. \ref{tab: inference-ablation}, the model can achieve similar fidelity and diversity with a slight decrease in instance quantity score.

\section{Conclusion}



This paper introduces a prompt-free conditional diffusion framework for multi-object image augmentation. Through the proposed local-global semantic fusion strategy and the reward model based counting loss, the model can augment the images in a large-scale and diverse manner that conforms to the original category distribution. Qualitative and quantitative experimental evaluations substantiate the efficacy and superiority of our proposed methodology. At the same time, both out-of-domain generalization ability and recurrent augmentation ability of the model provide more possibilities for its application.

\section*{Acknowledgments}
This work is supported in part by the National Natural Science Foundation of China under Grand 62372379, Grant 62472359, and Grant 62472350; in part by the Xi’an’s Key Industrial Chain Core Technology Breakthrough Project: AI Core Technology Breakthrough under Grand 23ZDCYJSGG0003-2023; in part by National Key Laboratory of Science and Technology on Space-Born Intelligent Information Processing fundation under Grant TJ-04-23-04; in part by Innovation Foundation for Doctor Dissertation of Northwestern Polytechnical University under Grant CX2025092.

\bibliographystyle{named}
\bibliography{ijcai25}

\begin{thebibliography}{}

\bibitem[\protect\citeauthoryear{Antoniou \bgroup \em et al.\egroup }{2017}]{GANaug}
Antreas Antoniou, Amos Storkey, and Harrison Edwards.
\newblock Data augmentation generative adversarial networks.
\newblock {\em arXiv preprint arXiv:1711.04340}, 2017.

\bibitem[\protect\citeauthoryear{Battash \bgroup \em et al.\egroup }{2024}]{T2I3}
Barak Battash, Amit Rozner, Lior Wolf, and Ofir Lindenbaum.
\newblock Obtaining favorable layouts for multiple object generation, 2024.

\bibitem[\protect\citeauthoryear{Binyamin \bgroup \em et al.\egroup }{2024}]{T2I1}
Lital Binyamin, Yoad Tewel, Hilit Segev, Eran Hirsch, Royi Rassin, and Gal Chechik.
\newblock Make it count: Text-to-image generation with an accurate number of objects, 2024.

\bibitem[\protect\citeauthoryear{Chen \bgroup \em et al.\egroup }{2023}]{generate4UDA}
Weijie Chen, Haoyu Wang, Shicai Yang, Lei Zhang, Wei Wei, Yanning Zhang, Luojun Lin, Di~Xie, and Yueting Zhuang.
\newblock Adapt anything: Tailor any image classifiers across domains and categories using text-to-image diffusion models, 2023.

\bibitem[\protect\citeauthoryear{Deng \bgroup \em et al.\egroup }{2009}]{ImageNet}
Jia Deng, Wei Dong, Richard Socher, Li-Jia Li, Kai Li, and Li~Fei-Fei.
\newblock Imagenet: A large-scale hierarchical image database.
\newblock In {\em 2009 IEEE conference on computer vision and pattern recognition}, pages 248--255. Ieee, 2009.

\bibitem[\protect\citeauthoryear{Dosovitskiy \bgroup \em et al.\egroup }{2020}]{vit}
Alexey Dosovitskiy, Lucas Beyer, Alexander Kolesnikov, Dirk Weissenborn, Xiaohua Zhai, Thomas Unterthiner, Mostafa Dehghani, Matthias Minderer, Georg Heigold, Sylvain Gelly, et~al.
\newblock An image is worth 16x16 words: Transformers for image recognition at scale.
\newblock {\em arXiv preprint arXiv:2010.11929}, 2020.

\bibitem[\protect\citeauthoryear{Gal \bgroup \em et al.\egroup }{2022}]{TextualInversion}
Rinon Gal, Yuval Alaluf, Yuval Atzmon, Or~Patashnik, Amit~H. Bermano, Gal Chechik, and Daniel Cohen-Or.
\newblock An image is worth one word: Personalizing text-to-image generation using textual inversion, 2022.

\bibitem[\protect\citeauthoryear{He \bgroup \em et al.\egroup }{2016}]{resnet}
Kaiming He, Xiangyu Zhang, Shaoqing Ren, and Jian Sun.
\newblock Deep residual learning for image recognition.
\newblock In {\em Proceedings of the IEEE conference on computer vision and pattern recognition}, pages 770--778, 2016.

\bibitem[\protect\citeauthoryear{He \bgroup \em et al.\egroup }{2017}]{maskrcnn}
Kaiming He, Georgia Gkioxari, Piotr Doll{\'a}r, and Ross Girshick.
\newblock Mask r-cnn.
\newblock In {\em Proceedings of the IEEE international conference on computer vision}, pages 2961--2969, 2017.

\bibitem[\protect\citeauthoryear{He \bgroup \em et al.\egroup }{2023}]{generate4classification}
Ruifei He, Shuyang Sun, Xin Yu, Chuhui Xue, Wenqing Zhang, Philip Torr, Song Bai, and XIAOJUAN QI.
\newblock Is synthetic data from generative models ready for image recognition?
\newblock In {\em The Eleventh International Conference on Learning Representations}, 2023.

\bibitem[\protect\citeauthoryear{Heusel \bgroup \em et al.\egroup }{2017}]{FrechetInceptionDistance}
Martin Heusel, Hubert Ramsauer, Thomas Unterthiner, Bernhard Nessler, and Sepp Hochreiter.
\newblock Gans trained by a two time-scale update rule converge to a local nash equilibrium.
\newblock {\em Advances in neural information processing systems}, 30, 2017.

\bibitem[\protect\citeauthoryear{Ho and Salimans}{2022}]{CFG}
Jonathan Ho and Tim Salimans.
\newblock Classifier-free diffusion guidance, 2022.

\bibitem[\protect\citeauthoryear{Hu \bgroup \em et al.\egroup }{2021}]{LoRA}
Edward~J. Hu, Yelong Shen, Phillip Wallis, Zeyuan Allen-Zhu, Yuanzhi Li, Shean Wang, Lu~Wang, and Weizhu Chen.
\newblock Lora: Low-rank adaptation of large language models, 2021.

\bibitem[\protect\citeauthoryear{Jocher \bgroup \em et al.\egroup }{2023}]{YOLOv8}
Glenn Jocher, Ayush Chaurasia, and Jing Qiu.
\newblock {Ultralytics YOLOv8}.
\newblock \url{https://github.com/ultralytics/ultralytics}, 2023.
\newblock Accessed: 2024-09-19.

\bibitem[\protect\citeauthoryear{Kirillov \bgroup \em et al.\egroup }{2023}]{SAM}
Alexander Kirillov, Eric Mintun, Nikhila Ravi, Hanzi Mao, Chloe Rolland, Laura Gustafson, Tete Xiao, Spencer Whitehead, Alexander~C Berg, Wan-Yen Lo, et~al.
\newblock Segment anything.
\newblock In {\em Proceedings of the IEEE/CVF international conference on computer vision}, pages 4015--4026, 2023.

\bibitem[\protect\citeauthoryear{Li \bgroup \em et al.\egroup }{2024}]{ControlNet++}
Ming Li, Taojiannan Yang, Huafeng Kuang, Jie Wu, Zhaoning Wang, Xuefeng Xiao, and Chen Chen.
\newblock Controlnet++: Improving conditional controls with efficient consistency feedback, 2024.

\bibitem[\protect\citeauthoryear{Lin \bgroup \em et al.\egroup }{2014}]{COCO}
Tsung-Yi Lin, Michael Maire, Serge Belongie, James Hays, Pietro Perona, Deva Ramanan, Piotr Doll{\'a}r, and C~Lawrence Zitnick.
\newblock Microsoft coco: Common objects in context.
\newblock In {\em Computer Vision--ECCV 2014: 13th European Conference, Zurich, Switzerland, September 6-12, 2014, Proceedings, Part V 13}, pages 740--755. Springer, 2014.

\bibitem[\protect\citeauthoryear{Liu \bgroup \em et al.\egroup }{2024}]{GroundingDINO}
Shilong Liu, Zhaoyang Zeng, Tianhe Ren, Feng Li, Hao Zhang, Jie Yang, Qing Jiang, Chunyuan Li, Jianwei Yang, Hang Su, Jun Zhu, and Lei Zhang.
\newblock Grounding dino: Marrying dino with grounded pre-training for open-set object detection, 2024.

\bibitem[\protect\citeauthoryear{Loshchilov and Hutter}{2019}]{AdamW}
Ilya Loshchilov and Frank Hutter.
\newblock Decoupled weight decay regularization, 2019.

\bibitem[\protect\citeauthoryear{Nguyen \bgroup \em et al.\egroup }{2023}]{DatasetDiffusion}
Quang Nguyen, Truong Vu, Anh Tran, and Khoi Nguyen.
\newblock Dataset diffusion: Diffusion-based synthetic data generation for pixel-level semantic segmentation.
\newblock In A.~Oh, T.~Naumann, A.~Globerson, K.~Saenko, M.~Hardt, and S.~Levine, editors, {\em Advances in Neural Information Processing Systems}, volume~36, pages 76872--76892. Curran Associates, Inc., 2023.

\bibitem[\protect\citeauthoryear{Nichol \bgroup \em et al.\egroup }{2022}]{GLIDE}
Alex Nichol, Prafulla Dhariwal, Aditya Ramesh, Pranav Shyam, Pamela Mishkin, Bob McGrew, Ilya Sutskever, and Mark Chen.
\newblock Glide: Towards photorealistic image generation and editing with text-guided diffusion models, 2022.

\bibitem[\protect\citeauthoryear{Podell \bgroup \em et al.\egroup }{2023}]{SDXL}
Dustin Podell, Zion English, Kyle Lacey, Andreas Blattmann, Tim Dockhorn, Jonas Müller, Joe Penna, and Robin Rombach.
\newblock Sdxl: Improving latent diffusion models for high-resolution image synthesis, 2023.

\bibitem[\protect\citeauthoryear{Radford \bgroup \em et al.\egroup }{2021}]{CLIP}
Alec Radford, Jong~Wook Kim, Chris Hallacy, Aditya Ramesh, Gabriel Goh, Sandhini Agarwal, Girish Sastry, Amanda Askell, Pamela Mishkin, Jack Clark, et~al.
\newblock Learning transferable visual models from natural language supervision.
\newblock In {\em International conference on machine learning}, pages 8748--8763. PMLR, 2021.

\bibitem[\protect\citeauthoryear{Ramesh \bgroup \em et al.\egroup }{2022}]{UnCLIP}
Aditya Ramesh, Prafulla Dhariwal, Alex Nichol, Casey Chu, and Mark Chen.
\newblock Hierarchical text-conditional image generation with clip latents, 2022.

\bibitem[\protect\citeauthoryear{Ravi \bgroup \em et al.\egroup }{2024}]{SAM2}
Nikhila Ravi, Valentin Gabeur, Yuan-Ting Hu, Ronghang Hu, Chaitanya Ryali, Tengyu Ma, Haitham Khedr, Roman Rädle, Chloe Rolland, Laura Gustafson, Eric Mintun, Junting Pan, Kalyan~Vasudev Alwala, Nicolas Carion, Chao-Yuan Wu, Ross Girshick, Piotr Dollár, and Christoph Feichtenhofer.
\newblock Sam 2: Segment anything in images and videos, 2024.

\bibitem[\protect\citeauthoryear{Rombach \bgroup \em et al.\egroup }{2022}]{StableDiffusion}
Robin Rombach, Andreas Blattmann, Dominik Lorenz, Patrick Esser, and Bj{\"o}rn Ommer.
\newblock High-resolution image synthesis with latent diffusion models.
\newblock In {\em Proceedings of the IEEE/CVF conference on computer vision and pattern recognition}, pages 10684--10695, 2022.

\bibitem[\protect\citeauthoryear{Ruiz \bgroup \em et al.\egroup }{2023}]{DreamBooth}
Nataniel Ruiz, Yuanzhen Li, Varun Jampani, Yael Pritch, Michael Rubinstein, and Kfir Aberman.
\newblock Dreambooth: Fine tuning text-to-image diffusion models for subject-driven generation.
\newblock In {\em Proceedings of the IEEE/CVF conference on computer vision and pattern recognition}, pages 22500--22510, 2023.

\bibitem[\protect\citeauthoryear{Saharia \bgroup \em et al.\egroup }{2022}]{Imagen}
Chitwan Saharia, William Chan, Saurabh Saxena, Lala Li, Jay Whang, Emily~L Denton, Kamyar Ghasemipour, Raphael Gontijo~Lopes, Burcu Karagol~Ayan, Tim Salimans, et~al.
\newblock Photorealistic text-to-image diffusion models with deep language understanding.
\newblock {\em Advances in neural information processing systems}, 35:36479--36494, 2022.

\bibitem[\protect\citeauthoryear{Suri \bgroup \em et al.\egroup }{2024}]{Gen2Det}
Saksham Suri, Fanyi Xiao, Animesh Sinha, Sean Culatana, Raghuraman Krishnamoorthi, Chenchen Zhu, and Abhinav Shrivastava.
\newblock Gen2det: Generate to detect.
\newblock In {\em Synthetic Data for Computer Vision Workshop@ CVPR 2024}, 2024.

\bibitem[\protect\citeauthoryear{Wang \bgroup \em et al.\egroup }{2024}]{InstanceDiffusion}
Xudong Wang, Trevor Darrell, Sai~Saketh Rambhatla, Rohit Girdhar, and Ishan Misra.
\newblock Instancediffusion: Instance-level control for image generation.
\newblock In {\em Proceedings of the IEEE/CVF Conference on Computer Vision and Pattern Recognition}, pages 6232--6242, 2024.

\bibitem[\protect\citeauthoryear{Wen \bgroup \em et al.\egroup }{2023}]{T2I2}
Song Wen, Guian Fang, Renrui Zhang, Peng Gao, Hao Dong, and Dimitris Metaxas.
\newblock Improving compositional text-to-image generation with large vision-language models, 2023.

\bibitem[\protect\citeauthoryear{Wu \bgroup \em et al.\egroup }{2023a}]{ParaDiffusion}
Weijia Wu, Zhuang Li, Yefei He, Mike~Zheng Shou, Chunhua Shen, Lele Cheng, Yan Li, Tingting Gao, Di~Zhang, and Zhongyuan Wang.
\newblock Paragraph-to-image generation with information-enriched diffusion model, 2023.

\bibitem[\protect\citeauthoryear{Wu \bgroup \em et al.\egroup }{2023b}]{DatasetDM}
Weijia Wu, Yuzhong Zhao, Hao Chen, Yuchao Gu, Rui Zhao, Yefei He, Hong Zhou, Mike~Zheng Shou, and Chunhua Shen.
\newblock Datasetdm: Synthesizing data with perception annotations using diffusion models.
\newblock {\em Advances in Neural Information Processing Systems}, 36:54683--54695, 2023.

\bibitem[\protect\citeauthoryear{Xie \bgroup \em et al.\egroup }{2023}]{MosaicFusion}
Jiahao Xie, Wei Li, Xiangtai Li, Ziwei Liu, Yew~Soon Ong, and Chen~Change Loy.
\newblock Mosaicfusion: Diffusion models as data augmenters for large vocabulary instance segmentation, 2023.

\bibitem[\protect\citeauthoryear{Xu \bgroup \em et al.\egroup }{2023}]{VersatileDiffusion}
Xingqian Xu, Zhangyang Wang, Gong Zhang, Kai Wang, and Humphrey Shi.
\newblock Versatile diffusion: Text, images and variations all in one diffusion model.
\newblock In {\em Proceedings of the IEEE/CVF International Conference on Computer Vision}, pages 7754--7765, 2023.

\bibitem[\protect\citeauthoryear{Xu \bgroup \em et al.\egroup }{2024}]{PromptFreeDiffusion}
Xingqian Xu, Jiayi Guo, Zhangyang Wang, Gao Huang, Irfan Essa, and Humphrey Shi.
\newblock Prompt-free diffusion: Taking" text" out of text-to-image diffusion models.
\newblock In {\em Proceedings of the IEEE/CVF Conference on Computer Vision and Pattern Recognition}, pages 8682--8692, 2024.

\bibitem[\protect\citeauthoryear{Yang \bgroup \em et al.\egroup }{2024}]{AddSD}
Lingfeng Yang, Xinyu Zhang, Xiang Li, Jinwen Chen, Kun Yao, Gang Zhang, Errui Ding, Lingqiao Liu, Jingdong Wang, and Jian Yang.
\newblock Add-sd: Rational generation without manual reference.
\newblock {\em arXiv preprint arXiv:2407.21016}, 2024.

\bibitem[\protect\citeauthoryear{Zhang \bgroup \em et al.\egroup }{2018}]{LPIPS}
Richard Zhang, Phillip Isola, Alexei~A Efros, Eli Shechtman, and Oliver Wang.
\newblock The unreasonable effectiveness of deep features as a perceptual metric.
\newblock In {\em Proceedings of the IEEE conference on computer vision and pattern recognition}, pages 586--595, 2018.

\bibitem[\protect\citeauthoryear{Zhang \bgroup \em et al.\egroup }{2023}]{ControlNet}
Lvmin Zhang, Anyi Rao, and Maneesh Agrawala.
\newblock Adding conditional control to text-to-image diffusion models.
\newblock In {\em Proceedings of the IEEE/CVF International Conference on Computer Vision}, pages 3836--3847, 2023.

\bibitem[\protect\citeauthoryear{Zhao \bgroup \em et al.\egroup }{2023}]{X-Paste}
Hanqing Zhao, Dianmo Sheng, Jianmin Bao, Dongdong Chen, Dong Chen, Fang Wen, Lu~Yuan, Ce~Liu, Wenbo Zhou, Qi~Chu, et~al.
\newblock X-paste: Revisiting scalable copy-paste for instance segmentation using clip and stablediffusion.
\newblock In {\em International Conference on Machine Learning}, pages 42098--42109. PMLR, 2023.

\end{thebibliography}


\begin{thebibliography}{}

\bibitem[\protect\citeauthoryear{Hu \bgroup \em et al.\egroup }{2021}]{LoRA}
Edward~J. Hu, Yelong Shen, Phillip Wallis, Zeyuan Allen-Zhu, Yuanzhi Li, Shean Wang, Lu~Wang, and Weizhu Chen.
\newblock Lora: Low-rank adaptation of large language models, 2021.

\bibitem[\protect\citeauthoryear{Liu \bgroup \em et al.\egroup }{2024}]{GroundingDINO}
Shilong Liu, Zhaoyang Zeng, Tianhe Ren, Feng Li, Hao Zhang, Jie Yang, Qing Jiang, Chunyuan Li, Jianwei Yang, Hang Su, Jun Zhu, and Lei Zhang.
\newblock Grounding dino: Marrying dino with grounded pre-training for open-set object detection, 2024.

\bibitem[\protect\citeauthoryear{Nguyen \bgroup \em et al.\egroup }{2023}]{DatasetDiffusion}
Quang Nguyen, Truong Vu, Anh Tran, and Khoi Nguyen.
\newblock Dataset diffusion: Diffusion-based synthetic data generation for pixel-level semantic segmentation.
\newblock In A.~Oh, T.~Naumann, A.~Globerson, K.~Saenko, M.~Hardt, and S.~Levine, editors, {\em Advances in Neural Information Processing Systems}, volume~36, pages 76872--76892. Curran Associates, Inc., 2023.

\bibitem[\protect\citeauthoryear{Ravi \bgroup \em et al.\egroup }{2024}]{SAM2}
Nikhila Ravi, Valentin Gabeur, Yuan-Ting Hu, Ronghang Hu, Chaitanya Ryali, Tengyu Ma, Haitham Khedr, Roman Rädle, Chloe Rolland, Laura Gustafson, Eric Mintun, Junting Pan, Kalyan~Vasudev Alwala, Nicolas Carion, Chao-Yuan Wu, Ross Girshick, Piotr Dollár, and Christoph Feichtenhofer.
\newblock Sam 2: Segment anything in images and videos, 2024.

\bibitem[\protect\citeauthoryear{Suri \bgroup \em et al.\egroup }{2024}]{Gen2Det}
Saksham Suri, Fanyi Xiao, Animesh Sinha, Sean Culatana, Raghuraman Krishnamoorthi, Chenchen Zhu, and Abhinav Shrivastava.
\newblock Gen2det: Generate to detect.
\newblock In {\em Synthetic Data for Computer Vision Workshop@ CVPR 2024}, 2024.

\bibitem[\protect\citeauthoryear{Szegedy \bgroup \em et al.\egroup }{2016}]{Inception}
Christian Szegedy, Vincent Vanhoucke, Sergey Ioffe, Jon Shlens, and Zbigniew Wojna.
\newblock Rethinking the inception architecture for computer vision.
\newblock In {\em Proceedings of the IEEE conference on computer vision and pattern recognition}, pages 2818--2826, 2016.

\bibitem[\protect\citeauthoryear{Xu \bgroup \em et al.\egroup }{2023}]{VersatileDiffusion}
Xingqian Xu, Zhangyang Wang, Gong Zhang, Kai Wang, and Humphrey Shi.
\newblock Versatile diffusion: Text, images and variations all in one diffusion model.
\newblock In {\em Proceedings of the IEEE/CVF International Conference on Computer Vision}, pages 7754--7765, 2023.

\bibitem[\protect\citeauthoryear{Zhang \bgroup \em et al.\egroup }{2018}]{LPIPS}
Richard Zhang, Phillip Isola, Alexei~A Efros, Eli Shechtman, and Oliver Wang.
\newblock The unreasonable effectiveness of deep features as a perceptual metric.
\newblock In {\em Proceedings of the IEEE conference on computer vision and pattern recognition}, pages 586--595, 2018.

\bibitem[\protect\citeauthoryear{Zheng \bgroup \em et al.\egroup }{2024}]{ZoneEval}
Zhaohui Zheng, Yuming Chen, Qibin Hou, Xiang Li, Ping Wang, and Ming-Ming Cheng.
\newblock Zone evaluation: Revealing spatial bias in object detection.
\newblock {\em IEEE Transactions on Pattern Analysis and Machine Intelligence}, 2024.

\end{thebibliography}

\end{document}


\maketitle

\section{Experiments Details}

\subsection{Detailed Implementation}

In our approach, we integrate the LoRA \cite{LoRA} into the UNet architecture, training it while keeping the other components of the network fixed. We describe in detail the hyperparameters used during training and inference in Tab. \ref{tab: train hyper-parameter}.

\begin{table}[htb]
\centering
\begin{tabular}{cc}
\toprule
\textbf{Name} & \textbf{Value} \\
\toprule
\textbf{Training} \\
Latent diffusion model & Stable Diffusion XL 1.0 \\
Reward model & Grounding DINO tiny \\
Input size & 512 $ \times $ 512 \\
Total batch size & 32 \\
Epoch & 5 \\
Augmentation & CenterCrop \& RandomFlip \\
Seed & 42 \\
Learning rate & 1e-4 \\
Learning rate scheduler & constant \\
Warmup steps & 0 \\
Optimizer & AdamW \\
AdamW $ \beta_1 $ & 0.9 \\
AdamW $ \beta_2 $ & 0.999 \\
Weight decay & 0.01 \\
Gradient clipping & 1.0 \\
Lora rank & 128 \\
Counting loss threshold & 0.1 \\
Counting loss weight & 0.5 \\
Counting loss steps & 1000 \\
Content length & 9 \\
Mixed Precision & FP16 \\
GPUs for Training & 2 $ \times $ NVIDIA 3090 \\
Training time & $ \sim $ 60 Hours \\
\midrule
\textbf{Inference} \\
Noise scheduler & EulerDiscreteScheduler \\
Inference steps & 50 \\
Guidance scale & 7.5 \\
height & 512 \\
width & 512 \\
\bottomrule
\end{tabular}
\caption{Training and inference hyperparameters.}
\label{tab: train hyper-parameter}
\end{table}

\subsection{Detailed Metrics}

In addition to the instance quantity score (IQS) we proposed, we also used two existing indicators to comprehensively evaluate the model. FID (Frechet Inception Distance score) is a metric used to evaluate the effect of generative models. It uses a pre-trained Inception \cite{Inception} network to convert real images and generated images into feature vector sets respectively. Then, the mean and covariance matrix of the two feature vector sets are calculated. Finally, the FID indicator is calculated by calculating the square root of the sum of the squares of the difference between the two covariance matrices. The diversity score (DS) is calculated using the LSIPS (Learned Perceptual Image Patch Similarity) \cite{LPIPS} mean of all images. LSIPS  is an indicator used to evaluate the perceptual similarity between images. It is a linear weighted distance metric based on deep network features. The calculation method is to feed the input image and augmented image into a pre-trained deep network, extract the output features of each layer, and normalize them. Then the features of each layer are linearly weighted, the L2 distance is calculated, and finally the LPIPS value is averaged.

\subsection{Detailed Compare Methods}

We compared our method with several existing approaches. For the method based on the text-to-image diffusion model, we selected DatasetDiffusion \cite{DatasetDiffusion}, a method based on category names as the comparison method. Specifically, since it has not been experimented on the COCO validation set, we used the official caption of the COCO validation set with the category prompt method it proposed, and using the negative prompts it set re-implemented it on SDXL and SDXL img2img. The method relying on mixed conditions is not ideal for dataset enhancement due to its inherent scalability issues, making it challenging to apply broadly. The method based on multiple single-object images will generate images that do not conform to the facts, resulting in a decrease in the effect of downstream tasks \cite{Gen2Det}. The method based on object replacement will deepen the spatial bias of the model because it does not change the background and layout of the image, resulting in a decrease in the effect of downstream tasks \cite{ZoneEval}. For image augmentation methods based on image variation, we selected a variety of state-of-the-art methods for experiments. We used its pre-trained model to perform image diversification on the COCO validation set.

\section{Further Analysis}

\subsection{Hyper-Parameter Sensitivity}

This framework involves some key hyperparameters, including the threshold parameter $ \tau $ in Eq. 6, the counting loss step $ \gamma $, and the weight parameter $ \lambda $ in Eq. 8. In order to demonstrate the effect of these hyperparameters, we use different settings of these hyperparameters to test the proposed method. As shown in Tab. \ref{tab: hyperparameter-ablation}, our method is robust to the setting of hyperparameters.

\begin{table}[htb]
\centering
\begin{subtable}[t]{0.9\linewidth}
\centering
\begin{tabular}{ccccc}
\toprule
$ \tau $ & \textbf{FID} $ \downarrow $ & \textbf{DS} $ \uparrow $ & \textbf{IQS} $ \uparrow $ & \textbf{IQS50} $ \uparrow $ \\
\midrule
0.1 & 18.59 & 0.71 & 29.17 & 34.15 \\
0.2 & 19.01 & 0.71 & 29.10 & 34.30 \\
0.3 & 19.06 & 0.71 & 28.60 & 33.51 \\
0.4 & 19.25 & 0.71 & 28.75 & 33.49 \\
0.5 & 19.43 & 0.71 & 29.09 & 33.94 \\
\bottomrule
\end{tabular}
\caption{Effect of $ \tau $}
\end{subtable}

\begin{subtable}[t]{0.9\linewidth}
\centering
\begin{tabular}{ccccc}
\toprule
$ \gamma $ & \textbf{FID} $ \downarrow $ & \textbf{DS} $ \uparrow $ & \textbf{IQS} $ \uparrow $ & \textbf{IQS50} $ \uparrow $ \\
\midrule
0 & 17.86 & 0.71 & 28.99 & 33.68 \\
1000 & 18.59 & 0.71 & 29.17 & 34.15 \\
3000 & 18.21 & 0.71 & 28.36 & 33.99 \\
5000 & 18.78 & 0.72 & 29.16 & 33.60 \\
10000 & 17.83 & 0.71 & 28.60 & 33.73 \\
15000 & 18.22 & 0.71 & 29.09 & 34.04 \\
\bottomrule
\end{tabular}
\caption{Effect of $ \gamma $}
\end{subtable}

\begin{subtable}[t]{0.9\linewidth}
\centering
\begin{tabular}{ccccc}
\toprule
$ \lambda $ & \textbf{FID} $ \downarrow $ & \textbf{DS} $ \uparrow $ & \textbf{IQS} $ \uparrow $ & \textbf{IQS50} $ \uparrow $ \\
\midrule
0.5 & 18.41 & 0.71 & 29.17 & 34.15 \\
1.0 & 18.50 & 0.71 & 28.45 & 33.21 \\
2.0 & 18.32 & 0.71 & 28.40 & 33.18 \\
\bottomrule
\end{tabular}
\caption{Effect of $ \lambda $}
\end{subtable}

\caption{Effect of different hyperparameters}
\label{tab: hyperparameter-ablation}
\end{table}

\subsection{Multiple Random Generatation \& Recurrent Generation}

\begin{figure*}[htbp]
    \centering
    \includegraphics[width=1.0\linewidth]{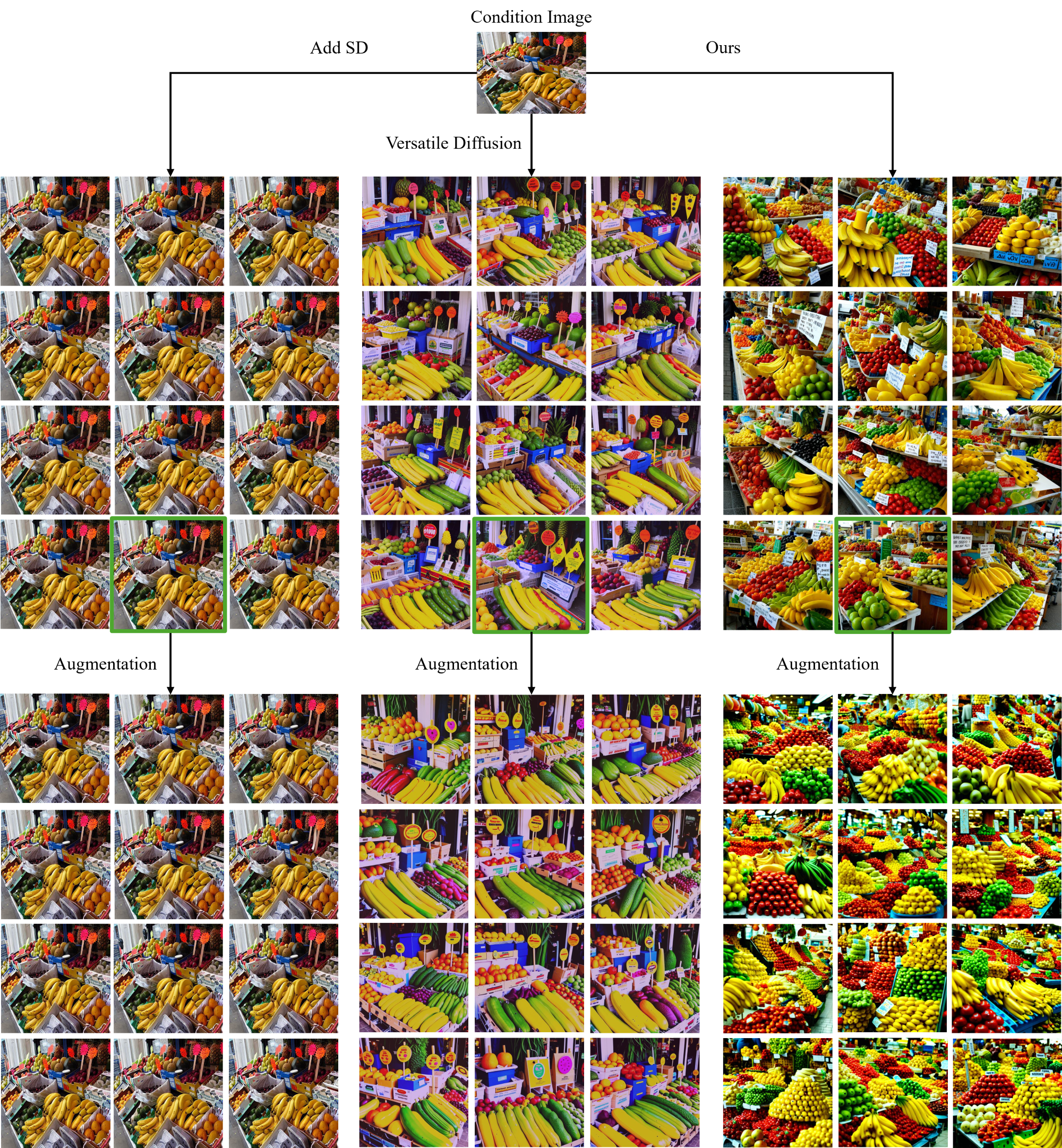}
    \caption{Recurrent generation for a given condition}
    \label{fig: variation1}
\end{figure*}

\begin{figure*}[htbp]
    \centering
    \includegraphics[width=1.0\linewidth]{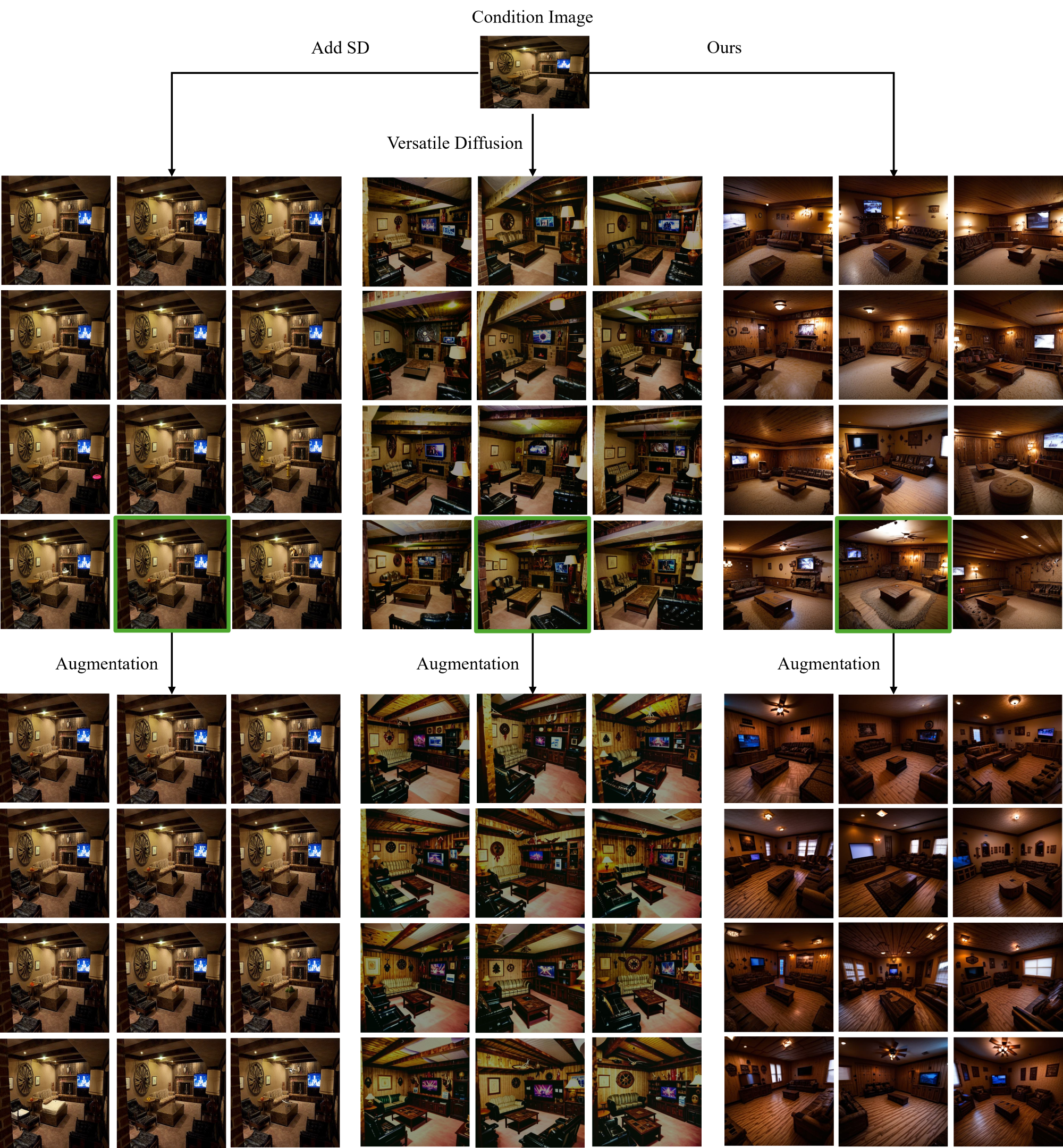}
    \caption{Recurrent generation for another given condition}
    \label{fig: variation2}
\end{figure*}

Dataset augmentation is not just a one-time augmentation of a single image, but also the diversity of multiple and recurrent augmentation of a single image. In order to verify the ability of the model, we compared the proposed method with the optimal instance quantity score comparison methods DatasetDiffusion \cite{DatasetDiffusion} w/SDXL img2img and VersatileDiffusion \cite{VersatileDiffusion}. As shown in Fig. \ref{fig: variation1} - \ref{fig: variation2}, we infer not only the diversified data augmentation of a single image but also the diversified generation based on the augmented image. The quantitative comparison results are shown in Table \ref{tab: large scale}. We calculated the standard deviations between the three color channels of 200 images, including 100 images based on the original images and 100 images based on the generated images, generated using different methods. Our method has a larger standard deviation, that is, better diversity.

\begin{table}[h]
\centering
\begin{tabular}{
>{\centering\arraybackslash}m{1.7cm}
>{\centering\arraybackslash}m{2.5cm}
>{\centering\arraybackslash}m{2.8cm}
}
\toprule
\textbf{Method} & \textbf{Average std of first generation} & \textbf{Average std of recurrent generation} \\
\toprule
\toprule
\textbf{Fig. \ref{fig: variation1}} \\
\midrule
Dataset Diffusion & 0.26 & 0.27 \\
\midrule
Versatile Diffusion & 0.27 & 0.27 \\
\midrule
Ours & \textbf{0.29} & \textbf{0.32} \\
\midrule
\midrule
\textbf{Fig. \ref{fig: variation2}} \\
\midrule
Dataset Diffusion & 0.18 & 0.18 \\
\midrule
Versatile Diffusion & 0.18 & 0.18 \\
\midrule
Ours & \textbf{0.19} & \textbf{0.20} \\
\bottomrule
\bottomrule
\end{tabular}
\caption{Average std for different methods}
\label{tab: large scale}
\end{table}

\subsection{Qualitive Ablation Result}

\begin{figure*}[htbp]
    \centering
    \includegraphics[width=0.7\linewidth]{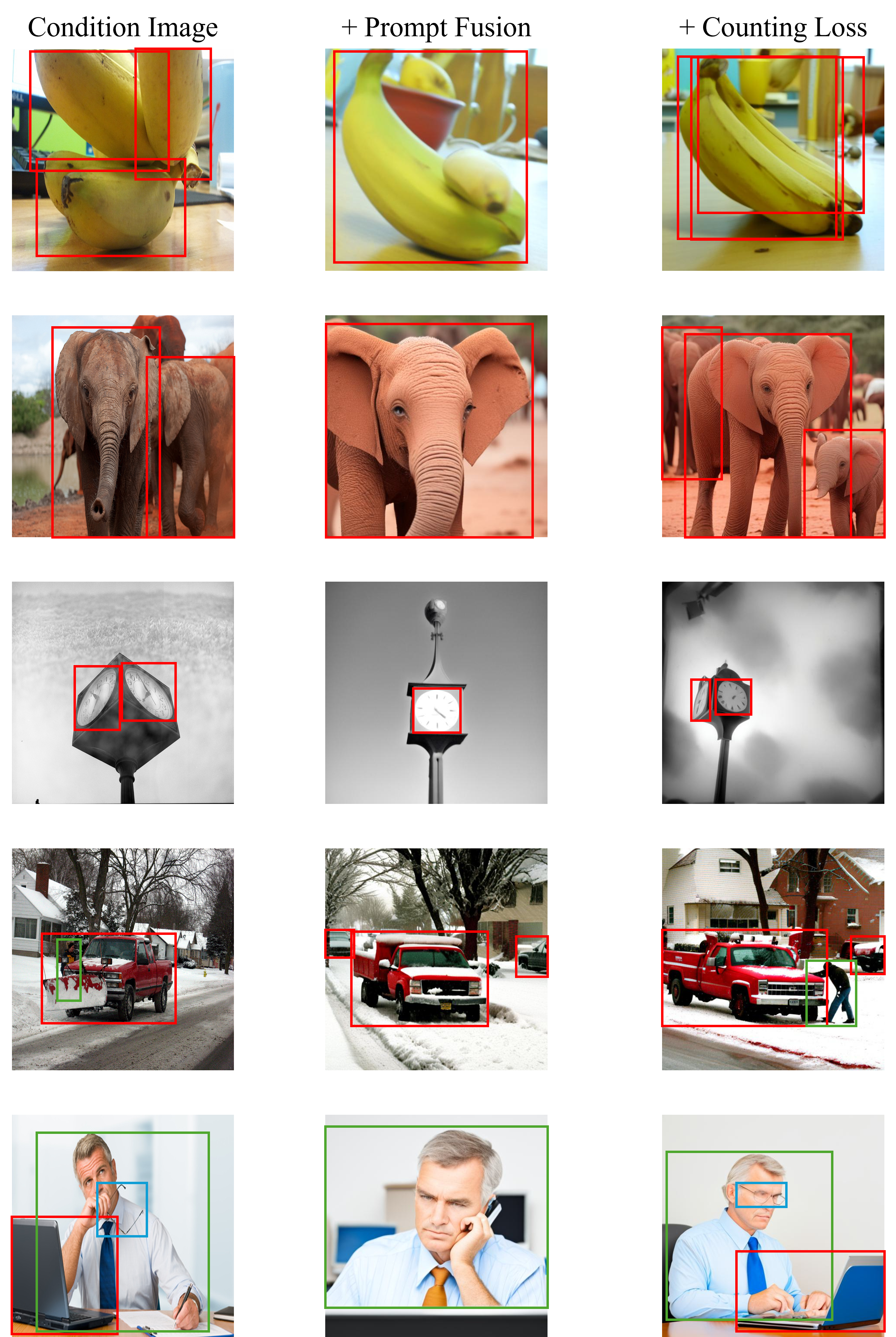}
    \caption{Qualitive ablation result. We manually annotated the instances for better visualization.}
    \label{fig: ablation}
\end{figure*}

To further analyze the role of each component, we also visualized the effect of adding the proposed methods one by one under the same seed setting. As shown in Fig. \ref{fig: ablation}, the local-global prompt fusion strategy can generate diverse images, but the object amount in each category of the images will be decreased, with problems of missing categories or reduced quantity. After adding the reward model based counting loss, diverse images with sufficient objects comparable to the original images can be generated.

\subsection{Challenges and Outlook}

There are several challenges left behind in this work:

\begin{itemize}
\item How to get the image annotations for various tasks after obtaining the image needs further exploration. We believe that with the continuous development of SAM \cite{SAM2} and Grounding DINO \cite{GroundingDINO}, image annotation will become simpler and simpler.
\item In addition to image annotation, our proposed method also has advanced out-of-domain image augmentation capabilities. Due to the constraints of our training data, the generated images primarily reflect natural scene perspectives. We will study the augmentation of more professional images such as medicine and SAR in future work.
\item Due to the limitation of training resources, we currently use pre-trained CLIP with LoRA to train the model, which limits the learning of image conditions. Moving forward, we aim to develop enhanced conditional encoding techniques to significantly boost the model's ability to augment images.
\end{itemize}

\bibliographystyle{named}
\bibliography{ijcai25}